\documentclass[runningheads,breaklinks,colorlinks]{llncs}
\usepackage[T1]{fontenc}
\usepackage{amsmath,amsfonts}
\usepackage{algorithmic}
\usepackage{array}
\usepackage{tabularx}
\usepackage{textcomp}
\usepackage{stfloats}
\usepackage{url}
\usepackage{verbatim}
\usepackage{graphicx}
\usepackage{cite}
\usepackage{balance}
\usepackage{graphicx}
\usepackage{subcaption}
\usepackage{dashrule}
\usepackage{wrapfig}

\usepackage{orcidlink}

\usepackage{booktabs}
\usepackage{amssymb} 
\usepackage{graphicx}
\usepackage{expl3}

\ExplSyntaxOn
\newcommand\latinabbrev[1]{
  \peek_meaning:NTF . {
    #1\@}%
  { \peek_catcode:NTF a {
      #1.\@ }%
    {#1.\@}}}
\ExplSyntaxOff

\newcommand{\method}{{SnapPose3D}}

\def\etal{\latinabbrev{et al}}

\begin{document}
\title{SnapPose3D: Diffusion-Based Single-Frame 2D-to-3D Lifting of Human Poses}
\titlerunning{Diffusion-Based Single-Frame 2D-to-3D Lifting of Human Poses}
%
\author{
Alessandro~Simoni\inst{1} \and
Riccardo~Catalini\inst{1} \and
Davide~Di~Nucci\inst{1} \and
Guido~Borghi\inst{1} \and
Davide~Davoli\inst{2}\thanks{Providing contracted services} \and
Lorenzo~Garattoni\inst{2} \and
Gianpiero~Francesca\inst{2} \and
Yuki~Kawana\inst{3} \and
Roberto~Vezzani\inst{1}
}

\authorrunning{Simoni et al.}

\institute{
University of Modena and Reggio Emilia (UniMoRe), Italy \and
Toyota Motor Europe (TME), Brussels, Belgium \and
Woven by Toyota, Japan
}
\maketitle 
\begin{abstract}
Depth ambiguity and joint uncertainty are the two main obstacles in obtaining accurate human pose predictions by 2D-to-3D lifting methods proposed in the literature.
In particular, these issues are caused by 2D joint locations that can be mapped to multiple 3D positions, inducing multiple possible final poses.
Following these considerations, we propose leveraging diffusion-based models' generation capability to predict multiple hypotheses and aggregate them in a final accurate pose.
Therefore, we introduce SnapPose3D, a pose-lifting framework trained deterministically to denoise 3D poses conditioned on both visual context and 2D pose features. SnapPose3D adopts a probabilistic approach during inference, generating multiple hypotheses through random sampling from a unit Gaussian distribution.
Unlike most previous methods that address pose ambiguity by processing temporal sequences, SnapPose3D uses single frames as input, avoiding tracking and limiting computational cost, data acquisition complexity, and the need for online, real-time applications.  
We extensively evaluate SnapPose3D on well-known benchmarks for the 3D human pose estimation task
showing its ability to generate and aggregate accurate hypotheses that lead to state-of-the-art results.
\end{abstract}

\begin{figure*}[t]
    \centering \includegraphics[width=1.0\linewidth]{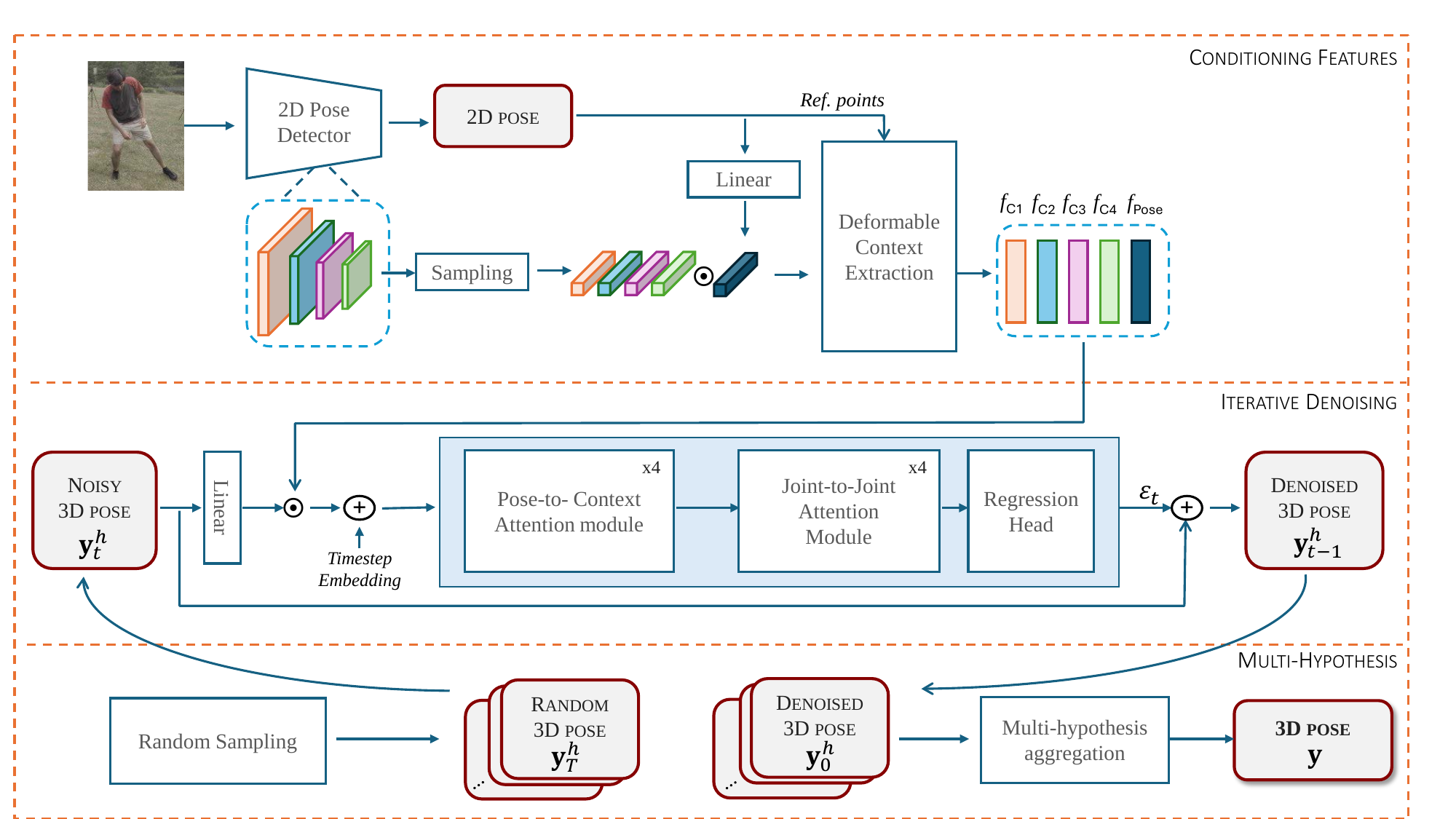}
    \caption{Overview of {\method} framework composed of three steps: (i) extraction of the conditioning features, \textit{i.e.} visual features computed started from the initial 2D pose and the context near the subject; (ii) then, a transformer-based architecture iteratively removes the noise $\epsilon_t$ from each input pose $\textbf{y}^h_t$; (iii) finally, the final pose is predicted by aggregating $H$ multiple denoised hypotheses.}
    \label{fig:model}
\end{figure*}

\section{Introduction}
\label{sec:intro}
The estimation of human poses from single images has been a topic of significant interest and importance for several years in the computer vision community, owing to its wide range of applications, including human-robot interaction~\cite{terreran2023general,simoni2022semi}, worker safety~\cite{paudel2022industrial,fan20243d}, and social behavior analysis~\cite{arac2019deepbehavior,liu2022ego+}.
However, while estimating the 2D joint positions on images is a well-defined problem and numerous methods already achieve extreme accuracy~\cite{sun2019deep,8765346,xu2022vitpose} on well-established benchmark datasets~\cite{li2019crowdpose,lin2014microsoft,mehta2017monocular}, the estimation of the 3D joint positions still poses significant challenges. 

The main issues are related to the depth ambiguity and joint location uncertainty, due to the inherent loss of three-dimensional information when projecting to monocular 2D. Besides, difficulties in collecting 3D data and annotations in realistic settings further increase the challenges~\cite{wang2021deep}. 
Despite these challenges, 3D Human Pose Estimation (3D HPE)~\cite{sarafianos20163d,simoni2024d} from single images is an important task and attracts significant interest in the computer vision community.

In this context, we introduce \textbf{SnapPose3D}, a diffusion-based framework that accurately estimates 3D human poses directly from a single input image. 

Following a common trend in the literature~\cite{pavllo20193d,zheng20213d,d2021refinet}, the 2D pose estimation procedure represents the starting point, due to the excellent performance of currently available methods~\cite{cao2017realtime,chen2018cascaded,sun2019deep}. Subsequently, the initial 2D pose is lifted to a set of multiple and plausible 3D poses through a conditioned diffusion-based mechanism~\cite{sohl2015deep}; these poses are then finally aggregated in a single 3D pose.

We show that the advantages of generating multiple hypotheses are twofold: firstly, their aggregation is a key element in solving the depth ambiguity since it allows for discarding outliers or less plausible predictions. 
Secondly, the analysis of the distribution of the hypotheses enables the inference of the degree of confidence in the predictions.
From a technical perspective, SnapPose3D is able to generate different hypotheses through a transformer-based architecture~\cite{vaswani2017attention}, leveraging the knowledge of conditioning features, \textit{i.e.} the initial 2D human pose and multi-resolution visual features extracted from the contest around the subject.
Additionally, SnapPose3D is based on a single frame as input: we deliberately refrain from incorporating temporal information as done by most of the previous works~\cite{gong2023diffpose,choi2023diffupose,shan2023diffusion},  then reducing the complexity of data acquisition and enabling online applications. 
Indeed, temporal data integration typically involves estimating a human pose across consecutive frames using, for instance, a sliding window or more complex techniques. However, such a processing approach is feasible only in offline scenarios where subsequent frames are readily available or where a degree of latency in result delivery is acceptable. Moreover, temporal tracking becomes imperative in multi-person scenarios. Conversely, estimating the 3D pose from a single frame can be done before tracking individuals.

We prove the efficacy of SnapPose3D through the evaluation of two well-known benchmark datasets for 3D human pose estimation, namely Human3.6M~\cite{ionescu2013human3} (including its hard subset H36MA~\cite{wehrbein2021probabilistic}) and MPI-INF-3DHP~\cite{mehta2017monocular}. 
The experimental results show that our method outperforms the performance of sota competitors.
In particular, we showcase a robust correlation between the reliability of the estimated pose and the similarity among the generated hypotheses. 
Finally, a qualitative evaluation of in-the-wild videos from 3DPW~\cite{von2018recovering} is reported.

\section{Related work}  
\subsection{3D Human Pose Estimation (3D HPE)} 

Most methods for 3D HPE are based on a preliminary 2D pose estimation phase followed by a lifting step that recovers the 3D original position starting from its 2D projection. Thanks to the high reliability of state-of-the-art 2D pose estimators \cite{chen2018cascaded,sun2019deep}, the retrieval of the missing third coordinate is the most challenging part of 3D HPE.  
Without any additional information, the 2D-to-3D lifting is almost an ill-posed problem. Thus, temporal information \cite{pavllo20193d,zheng20213d,li2022mhformer}, available if processing a sequence of 2D postures at a time, or visual features of the person and the corresponding context are required to inject more constraints. 

The main limitations of the first class of methods are related to the latency generated by estimating the output of the central frame within a window and the need for tracking in case of multi-person scenarios. 
On the contrary, the second class of methods requires more computational load to reprocess the appearance features, as already done in the preliminary 2D pose estimation step.

As confirmed in~\cite{sminchisescu2003kinematic,simo2012single,d2023depth}, the variability of human postures does not allow the application of sufficient constraints to back-project 2D joints into the original 3D space. However, multiple 
hypotheses could be formulated to simultaneously satisfy the alignment to the 2D pose and additional conditions, such as anatomical constraints. 
For instance, Jahangiri \etal \cite{jahangiri2017generating} apply anatomical constraints such as joint angle and bone length to limit the number of hypotheses.
Li \etal \cite{li2022mhformer} proposes a multi-hypothesis Transformer to predict a 3D pose of a central frame, but it requires a long input sequence of 2D keypoints as additional temporal information.
Li \etal~\cite{li2019generating} combines a mixture density network \cite{bishop1994mixture} with the inverse problem of monocular 3D-HPE to generate plausible 3D poses.
Oikarinen \etal~\cite{oikarinen2021graphmdn} improves the mixture density network-based method~\cite{li2019generating} with the strength of semantic graph neural networks~\cite{zhaoCVPR19semantic}.

Conditional variational autoencoders have been exploited by
Sharma \etal \cite{sharma2019monocular} to sample 3D-pose candidates that are aggregated with a specific deep network to obtain the final prediction.
Wehrbein \etal \cite{wehrbein2021probabilistic} model the posterior distribution of 3D poses with normalizing flow by explicitly incorporating the uncertainty provided by the detector together with the 2D keypoint positions.

\subsection{Diffusion Models for 3D HPE} 
Following the impressive results obtained by Denoising Diffusion Probabilistic Models (DDPM) \cite{ho2020denoising}
in image synthesis  \cite{rombach2022high,dhariwal2021diffusion,song2021denoising}, the same approach has been applied to the 3D-HPE problem. 
 
In the HPE case, the human pose is considered a non-equilibrium system that diffuses to a completely random set of joint coordinates. Given this assumption, the DiffPose framework \cite{gong2023diffpose} considers the 3D pose estimation as a reverse diffusion process, where a 3D pose distribution with high uncertainty and indeterminacy is progressively transformed toward a 3D pose with low uncertainty. To guide the diffusion model, DiffPose leverages the spatial-temporal context extracted from a 2D pose sequence. However, a single-person video or a tracking algorithm is required. 
 
A key advantage of DDPMs for pose estimation is the natural generation of multiple hypotheses. A concurrent diffusion-based method, also named DiffPose~\cite{Holmquist_2023_ICCV}, conditions the process on 2D keypoint heatmaps from a single frame, limiting its applicability to heatmap-based methods. D3DP~\cite{shan2023diffusion} exploits this probabilistic nature to enhance performance through hypothesis aggregation.

A joint-wise reprojection-based multi-hypothesis aggregation (JPMA) method is proposed for limiting the influence of outliers when the intrinsic calibration parameters of the cameras are available. As the human pose has tight semantic connectivity between adjacent joints, Choi \etal \cite{choi2023diffupose} adopt a graph convolutional neural network (GCN) \cite{kipf2016semi} as a denoiser architecture to explicitly learn correlations between joints.

\section{SnapPose3D}
Given an image $I^{H \times W \times 3}$ as input, the goal of SnapPose is to predict the 3D human pose 
$\textbf{y} \in \mathbb{R}^{J \times 3}$ where $J$ is the number of skeleton's joints.

As visually summarized in Figure~\ref{fig:model}, the diffusion process is conditioned with features embedding the visual context near the subject -- which is usually discarded by temporal pose-lifting architectures -- and the 2D pose. Both of them are computed on the single input image.

\subsection{Preliminary}
Diffusion models~\cite{sohl2015deep} represent probabilistic generative approaches that learn to synthesize data through a parametrized Markov chain with variational inference. In particular, a diffusion model is composed of two recurrent processes: (i) \textit{forward diffusion} that creates noisy samples starting from a target data $x_0$, and (ii) \textit{reverse diffusion} that progressively denoises a random Gaussian noise to get the sample $x_0$. In the following, we follow the formulation of  Denoising Diffusion Probabilistic Models (DDPM)~\cite{ho2020denoising}.
To learn the reverse diffusion process, a set of intermediate noisy samples is needed to bridge the gap between the target sample and the Gaussian noise. 
The forward process is applied where the posterior distribution $q(x_{1:T}~|~x_0)$ is computed by gradually adding infinitesimal Gaussian noise $\epsilon$ with a variance $\beta_t \in [0, 1]$ for each timestep $t \in [0, T]$:

\begin{align}
q(x_{1:T}~|~x_0) := \prod_{t=1}^{T} q(x_t~|~x_{t-1}),  &\quad  q(x_t~|~x_{t-1}) := \mathcal{N}_{pdf}(x_t~|~\sqrt{1 - \beta_t}~x_{t-1}~,~\beta_t\mathbf{I})
\label{eq:forward_diffusion}
\end{align}

\noindent where $\mathcal{N}_{pdf}$ is the likelihood of sampling $x_t$ given the conditioning parameters, $T$ is the total number of diffusion steps, and $\mathbf{I}$ denotes the identity matrix. 

On the other hand, through the reverse process, the goal is to estimate the posterior $q(x_{t-1}~|~x_t)$ and thus find the target data $x_0$ or its distribution. Given $\alpha := 1 - \beta_t$ and $\bar{\alpha}_t := \prod^{t}_{k=1}\alpha_k$, we can apply the Bayes theorem and formulate $q(x_{t-1}~|~x_t)$ as follows:

\begin{align}
\tilde{\beta}_t := \frac{1 - \bar{\alpha}_{t-1}}{1 - \bar{\alpha}_t}\beta_t  &\qquad  \tilde{\mu}_t(x_t,x_0) := \frac{\sqrt{\bar{\alpha}_{t-1}}\beta_t}{1 - \bar{\alpha}_t}x_0 + \frac{\sqrt{\alpha_t}(1 - \bar{\alpha}_{t-1})}{1 - \bar{\alpha}_t}x_t
\end{align}



\begin{equation}
    q(x_{t-1}~|~x_t,x_0) = \mathcal{N}_{pdf}(\tilde{\mu}_t(x_t,t),\tilde{\beta}_t\mathbf{I})
\end{equation}

\noindent Since the above formulation needs to know $x_0$ in advance, diffusion models leverage neural networks to estimate the value of the posterior 
    $p_\theta(x_{t-1}~|~x_t) \simeq q(x_{t-1}~|~x_t,x_0)$.
\noindent Thus, a diffusion model aims to approximate the reverse process through a neural network learning the infinitesimal noise $\epsilon(x_t,t)$ between two consecutive timesteps. The objective loss function can be defined as
    $\mathcal{L} = \mathbb{E}_{x_0,t,\epsilon}[\parallel\epsilon - \epsilon_\theta(x_t,t)\parallel^2]$.

\subsection{Conditioning Features extraction}
\label{sec:features}

The conditioning of the reverse denoising is obtained with a set of features obtained by merging together appearance and 2D pose information. 
Firstly,  HRNet-32~\cite{sun2019deep} is used  
as a multi-resolution feature extractor to generate $f'_C = f'_{C1} \odot f'_{C2} \odot f'_{C3} \odot f'_{C4}$, where $\odot$ is the concatenation operator and $f'_{C1} \ldots f'_{C4}$ are hierarchical feature vectors (from $L=4$ levels) converted to a common size through sampling. In addition, HRNet-32 also generates the 2D pose, projected by a linear layer to $f'_P$ and concatenated to the visual features. 
Eventually, a pretrained Deformable Context Extraction module \cite{zhao2023contextaware} generates the final multichannel descriptor $F \in \mathbb{R}^{(L+1) \times J \times d}$, where $L=4$ is the number of feature levels, increased by one with the pose channel, $J$ is the number of joints and $d$ is the embedding size ($d=128$ in our experiments).

\subsection{Iterative Denoising procedure}
\label{sec:snappose3d}
The denoising network is a transformer-based network that takes as input the embedding previously defined and concatenated to a projection of the current 3D noisy pose. 
A timestep embedding $f_t$ is added to the concatenated features to make the model aware of the current diffusion step. 
Formally, the noisy pose $\textbf{y}_t \in \mathbb{R}^{J \times 3}$ is linearly projected into a higher-dimensional embedding of dimension $d=128$. The resulting input embedding is processed by two transformer-based modules that perform self-attention between pose and context and between each joint, respectively. Lastly, a regression head outputs the predicted noise $\hat{\epsilon}$ that was initially added to the ground truth pose $\textbf{y}_0$.

The Pose-to-Context Attention Module considers each channel as a token of size $d \times J$, in which $d$ is the size of the embedding. The Joint-to-Joint Attention Module considers each joint as a token of size $c \times d$, where $c$ is the total number of channels.  
 
The first module enriches each joint with contextual information, while the second one allows data sharing between the different joints.

We note that to apply the diffusion process to the 3D human pose estimation task, previous works~\cite{gong2023diffpose,shan2023diffusion,choi2023diffupose} took inspiration from the 2D-to-3D lifting literature, adapting state-of-the-art architectures as a denoiser function. In particular, existing lifting approaches~\cite{zheng20213d,li2022mhformer} predict the 3D pose of the central frame from a temporal sequence of 2D poses. However, this technique defines a \textit{non-causal} system requiring access also to future poses, that are not available in online real-world applications. In addition, a tracking system is required in multi-person scenarios, and the output is given with a non-negligible latency.

\subsection{Training and Inference}
During training, a timestep $t \in [0, T]$ is randomly selected for each sample in the batch. Subsequently, the forward diffusion process is applied to the ground truth 3D pose $\textbf{y}_0$ to obtain the noisy pose $\textbf{y}_t$. Following the definition in Equation~\ref{eq:forward_diffusion}, this operation can be formulated as
%
    $q(\textbf{y}_t~|~\textbf{y}_0) := \prod_{t} q(\textbf{y}_t~|~\textbf{y}_{t-1})$.
%
\noindent The resulting noisy pose $\textbf{y}_t$ is given as input to our denoiser network $\mathcal{D}$ conditioned on context features $f_C$, 2D pose features $f_P$ and timestep embedding $f_t$ to predict the added noise $\hat{\epsilon}$:
%
    $\hat{\epsilon} = \mathcal{D}(\textbf{y}_t, f_C, f_P, f_t)$.
%
\noindent The objective loss of the proposed method is an MSE loss computed on the noise as follows:
%
    $\mathcal{L} = \mathbb{E}_{\textbf{y}_0,t,\epsilon\sim\mathcal{N}(0,\mathbf{I})}[\parallel\epsilon - \hat{\epsilon}\parallel^2]$

\label{subsec:inference}
Since the corrupted pose $\textbf{y}_t$ with $t \rightarrow T$ can be approximated by a Gaussian distribution, the reverse process can start from $H$ initial hypotheses $\textbf{y}_{T}^h, h \in [1,H]$, each one obtained by sampling random noise from a unit Gaussian. 
These pure-noisy hypotheses are given as input to the denoiser $\mathcal{D}$ that, conditioned on the visual features extracted from the current frame, outputs the quantity of noise $\hat{\epsilon}_t$ which should be removed to get the 3D pose predictions $\textbf{y}_{t-1}^h$ at the next timestep.
During inference, we use an optimized version of DDPM denoising strategy, namely DDIM~\cite{song2021denoising}. For each hypothesis in parallel, the predicted noise $\hat{\epsilon}$ is removed from the corrupted pose as follows
%
    $\textbf{y}_{t-1}^h = \textbf{y}_{t}^h - \hat{\epsilon}_t$.
%
%
\noindent Starting with $t=T$, this procedure is iteratively repeated K times  (K=20 in our experiments) so that the timestep for each iteration is defined as $t = T \cdot (1 - k)/\text{K}, k \in [0, \text{K})$. 

\subsection{Multiple Hypothesis Aggregation} \label{subsec:aggregation}
As mentioned, the advantage of using a diffusion-based approach like \method~is that it combines the simplicity of deterministic training with the flexibility of probabilistic inference. 
Indeed, during training, SnapPose3D produces a single prediction while learning to handle Gaussian noise added to each sample. However, during inference, it generates multiple hypotheses by sampling from a Gaussian distribution, allowing an arbitrary number $H$ of predictions.

Following this approach, it is possible to output multiple plausible poses conditioned on the same input frame.  
If the intrinsic camera parameters are given, the re-projection error of the 3D poses when compared with the initial 2D poses could be exploited \cite{shan2023diffusion} as a selection schema, even if we prefer to ignore them, even when available, as in the most general case.
A single hypothesis can be randomly selected as the final pose estimation among the available $H$ ones (called \textbf{R}andom selection). The output pose can also be obtained by selecting the best hypothesis, \textit{i.e.} the one closest to the ground truth (\textbf{B}est selection). This approach is also referred to as ``Supervision from an Oracle''~\cite{sharma2019monocular}, but it is important to note that it requires the knowledge of the ground truth. 

However, generating multiple hypotheses not only increases the probability of obtaining at least one good solution but also produces additional information that can be exploited through proper hypothesis aggregation.

Given $H$ hypotheses $\textbf{y}_{0}^h$, a unique solution $\textbf{y}_{0}$ must be produced as output. Averaging (\textbf{A}) is the standard aggregation approach, though it is sensitive to outliers. To address this, we also employ the Median operator (\textbf{M}), which selects the central element after independently sorting the $3\times J$ vectors, each containing the $H$ hypotheses for each joint coordinate ($X,Y,Z$).

\begin{table*}[t!]
\centering
\caption{
Results in terms of MPJPE on Human3.6M using CPN~\cite{chen2018cascaded} 2D pose detections as input. Competitors are subdivided into three groups from top to bottom: (i) baselines (ii) diffusion approaches providing a single prediction through aggregation, (iii) diffusion approaches that select one of the hypotheses as output. For each part, best results are in bold, second ones are underlined. Number of hypotheses H and aggregation (\textbf{A}verage, \textbf{M}edian) or selection (\textbf{R}andom, \textbf{B}est) method are also reported.
}
\label{tab:mpjpe-human}
\resizebox{\linewidth}{!}{
\begin{tabular}{r | c c c c c c c c c c c c c c c c}
\toprule
Protocol 1 (\textbf{MPJPE}) (mm) & Dir & Dis & Eat & Gre & Phn & Pht & Pos & Pur & Sit & SitD & Smk & Wait & WlkD & Walk & WlkT & \textbf{Avg} \\
\midrule
\multicolumn{3}{l}{\textbf{Baseline methods}}& \multicolumn{14}{l}{\textbf{}} \\ \midrule
Pavlakos~\cite{pavlakos2017coarse} & 67.4 & 71.9 & 66.7 & 69.1 & 72.0 & 77.0 & 65.0 & 68.3 & 83.7 & 96.5 & 71.7 & 65.8 & 74.9 & 59.1 & 63.2 & 71.9 \\
Martinez~\cite{martinez2017simple} & 51.8 & 56.2 & 58.1 & 59.0 & 69.5 & 78.4 & 55.2 & 58.1 & 74.0 & 94.6 & 62.3 & 59.1 & 65.1 & 49.5 & 52.4 & 62.9 \\
Zhao~\cite{zhaoCVPR19semantic} & 48.2 & 60.8 & 51.8 & 64.0 & 64.6 & 53.6 & 51.1 & 67.4 & 88.7 & \underline{57.7} & 73.2 & 65.6 & 48.9 & 64.8 & 51.9 & 60.8 \\
Sun~\cite{sun2017compositional} & 52.8 & 54.8 & 54.2 & 54.3 & 61.8 & \underline{53.1} & 53.6 & 71.7 & 86.7 & 61.5 & 67.2 & 53.4 & 47.1 & 61.6 & 53.4 & 59.1 \\
Yang~\cite{yang20183d} & 51.5 & 58.9 & 50.4 & 57.0 & 62.1 & 65.4 & 49.8 & 52.7 & 69.2 & 85.2 & 57.4 & 58.4 & \underline{43.6} & 60.1 & 47.7 & 58.6 \\
Hossain~\cite{hossain2018exploiting} & 48.4 & 50.7 & 57.2 & 55.2 & 63.1 & 72.6 & 53.0 & 51.7 & 66.1 & 80.9 & 59.0 & 57.3 & 62.4 & 46.6 & 49.6 & 58.3 \\
Liu~\cite{liu2020comprehensive} & 46.3 & 52.2 & \underline{47.3} & 50.7 & 55.5 & 67.1 & 49.2 & \textbf{46.0} & 60.4 & 71.1 & 51.5 & 50.1 & 54.5 & 40.3 & 43.7 & 52.4 \\
Xu~\cite{xu2021graph} & \underline{45.2} & \underline{49.9} & 47.5 & 50.9 & \underline{54.9} & 66.1 & 48.5 & \underline{46.3} & \textbf{59.7} & 71.5 & \underline{51.4} & \underline{48.6} & 53.9 & \underline{39.9} & 44.1 & 51.9 \\
Zhao~\cite{zhao2022graformer} & \underline{45.2} & 50.8 & 48.0 & \underline{50.0} & \underline{54.9} & 65.0 & \underline{48.2} & 47.1 & \underline{60.2} & 70.0 & 51.6 & 48.7 & 54.1 & \textbf{39.7} & \underline{43.1} & \underline{51.8} \\
Zhao~\cite{zhao2023contextaware} & \textbf{37.7} & \textbf{42.0} & \textbf{39.0} & \textbf{42.7} & \textbf{43.5} & \textbf{37.6} & \textbf{41.1} & 51.4 & 61.7 & \textbf{43.4} & \textbf{49.4} & \textbf{43.0} & \textbf{33.9} & 45.2 & \textbf{35.0} & \textbf{43.4} \\
\midrule
\multicolumn{1}{l}{\textbf{Diffusion-based methods}} &  \multicolumn{16}{l}{\textbf{Aggregation of multiple hypotheses}} \\ \midrule
(H=10,A) Diffupose~\cite{choi2023diffupose} & 43.4 & 50.7 & 45.4 & 50.2 & 49.6 & 53.4 & 48.6 & \textbf{45.0} & \underline{56.9} & 70.7 & \underline{47.8} & 48.2 & 51.3 & 43.1 & 43.4 & 49.4 \\
(H=5,A) Diffpose~\cite{gong2023diffpose} & \underline{42.8} & 49.1 & 45.2 & 48.7 & 52.1 & 63.5 & 46.3 & \underline{45.2} & 58.6 & 66.3 & 50.4 & 47.6 & 52.0 & \textbf{37.6} & 40.2 & 49.7 \\
\multicolumn{17}{c}{\hdashrule[0.2ex]{17cm}{0.5pt}{1mm}} \\
%
(H=20, A) \textit{\method} & \textbf{36.4} & \underline{41.8} & \textbf{39.6} & \underline{42.1} & \underline{42.7} & \textbf{36.7} & \textbf{43.0} & 53.5 & 62.0 & \textbf{42.1} & 51.3 & \underline{40.6} & \textbf{32.0} & 45.3 & \underline{35.3} & \underline{43.2} \\
%
(H=20, M) \textit{\method} &
43.2 &
\textbf{41.3} &
\underline{41.4} &
\textbf{39.0} &
\textbf{40.5} &
\underline{44.8} &
\underline{43.9} &
47.5 &
\textbf{46.2} &
\underline{50.5} &
\textbf{43.2} &
\textbf{40.0} &
\underline{41.5} &
\underline{40.5} &
\textbf{34.5} &
\textbf{42.8} \\

\midrule
\multicolumn{1}{l}{\textbf{Diffusion-based methods}} & \multicolumn{16}{l}{\textbf{One hypothesis selected}} \\ \midrule
(H=200, B) DiffPose~\cite{Holmquist_2023_ICCV} &
 38.1 & 43.1 &\textbf{35.3} &43.1& 46.6& 48.2& 39.0& \textbf{37.6}& 51.9& 59.3& 41.7& 47.6& 45.4& 37.4& 36.0 & \underline{43.3}\\
\multicolumn{17}{c}{\hdashrule[0.2ex]{17cm}{0.5pt}{1mm}} \\

(H=20, R) \textit{\method} & \underline{38.7} & \underline{44.4} & \underline{41.6} & \underline{44.6} & \underline{45.0} & \underline{38.8} & \underline{45.6} & 55.6 & 65.3 &\underline{44.4} & 54.1 & \underline{43.1} & \underline{34.3} & 48.2 & \underline{37.7} & 45.6 \\

(H=20, B) \textit{\method} &
\textbf{37.5} &
\textbf{35.5} &
35.7 & 
\textbf{33.4} &
\textbf{34.9} &
\textbf{38.6} &
\textbf{37.8} &
41.1 &
\textbf{40.0} &
\textbf{43.7} &
\textbf{37.4} &
\textbf{34.5} &
\textbf{35.8} &
\textbf{34.8} &
\textbf{29.2} &
\textbf{36.9}
\\
\bottomrule
\end{tabular}
}
\end{table*}
\begin{table*}[t]
\centering
\caption{Results in terms of P-MPJPE on Human3.6M using CPN~\cite{chen2018cascaded} 2D pose detections as input. Further details are reported in the caption of  Table 1.
}
\label{tab:p-mpjpe-human}

\resizebox{\linewidth}{!}{
\begin{tabular}{r | c c c c c c c c c c c c c c c c}
\toprule
Protocol 2 (\textbf{P-MPJPE}) (mm) & Dir & Dis & Eat & Gre & Phn & Pht & Pos & Pur & Sit & SitD & Smk & Wait & WlkD & Walk & WlkT & \textbf{Avg} \\
\midrule
\multicolumn{3}{l}{
\textbf{Baseline methods}} \\ \midrule
Sun~\cite{sun2017compositional} & 42.1 & 44.3 & 45.0 & 45.4 & 51.5 & 53.0 & 43.2 & 41.3 & 59.3 & 73.3 & 51.0 & 44.0 & 48.0 & 38.3 & 44.8 & 48.3 \\
Martinez~\cite{martinez2017simple} & 39.5 & 43.2 & 46.4 & 47.0 & 51.0 & 56.0 & 41.4 & 40.6 & 56.5 & 69.4 & 49.2 & 45.0 & 49.5 & 38.0 & 43.1 & 47.7 \\
Hossain~\cite{hossain2018exploiting} & 36.9 & 37.9 & 42.8 & 40.3 & 46.8 & 46.7 & 37.7 & 36.5 & 48.9 & \underline{52.6} & 45.6 & 39.6 & 43.5 & 35.2 & 38.5 & 42.0 \\
Pavlakos~\cite{pavlakos2017coarse} & 34.7 & 39.8 & 41.8 & 38.6 & \underline{42.5} & 47.5 & 38.0 & 36.6 & 50.7 & 56.8 & 42.6 & 39.6 & 43.9 & \underline{32.1} & 36.5 & 41.8 \\
Liu~\cite{liu2020comprehensive} & 35.9 & 40.0 & 38.0 & 41.5 & \underline{42.5} & 51.4 & 37.8 & \underline{36.0} & \underline{48.6} & 56.6 & 41.8 & 38.3 & 42.7 & \textbf{31.7} & 36.2 & 41.2 \\
Yang~\cite{yang20183d} & \textbf{26.9} & \textbf{30.9} & \underline{36.3} & \underline{39.9} & 43.9 & \underline{47.4} & \textbf{28.8} & \textbf{29.4} & \textbf{36.9} & 58.4 & \underline{41.5} & \textbf{30.5} & \underline{29.5} & 42.5 & \underline{32.2} & \underline{37.7} \\
Zhao~\cite{zhao2023contextaware} & \underline{32.3} & \underline{34.6} & \textbf{33.1} & \textbf{35.1} & \textbf{35.2} & \textbf{30.1} & \underline{32.1} & 42.6 & 48.8 & \textbf{36.0} & \textbf{39.1} & \underline{33.1} & \textbf{27.5} & 37.1 & \textbf{29.3} & \textbf{35.4} \\
\midrule
\multicolumn{1}{l}{
\textbf{Diffusion-based methods} } &
\multicolumn{16}{l}{
\textbf{Aggregation of multiple hypotheses} }\\ \midrule
(H=10,A) Diffupose~\cite{choi2023diffupose} & 35.9 & 40.3 & 36.7 & 41.4 & 39.8 & 43.4 & 37.1 & \underline{35.5} & 46.2 & 59.7 & 39.9 & 38.0 & 41.9 & 32.9 & 34.2 & 39.9 \\
(H=5, A) Diffpose~\cite{gong2023diffpose} & \underline{33.9} & 38.2 & 36.0 & 39.2 & 40.2 & 46.5 & 35.8 & \textbf{34.8} & 48.0 & 52.5 & 41.2 & 36.5 & 40.9 & \textbf{30.3} & 33.8 & 39.2 \\

\multicolumn{17}{c}{\hdashrule[0.2ex]{17cm}{0.5pt}{1mm}} \\

(H=20, A) \textit{\method} & \textbf{31.5} & \underline{34.7} & \textbf{32.7} & \underline{35.1} & \underline{34.2} & \textbf{29.9} & \textbf{33.1} & 42.8 & \underline{47.5} & \textbf{34.4} & \underline{38.6} & \underline{31.6} & \textbf{26.5} & 36.7 & \underline{29.7} & \underline{34.8} \\

(H=20, M) \textit{\method} & 
35.7 &
\textbf{34.3} &
\underline{34.0} &
\textbf{32.5}&
\textbf{32.9}&
\underline{36.7}&
\underline{35.6}&
38.2&
\textbf{37.1}&
\underline{38.4}&
\textbf{34.3}&
\textbf{31.5}&
\underline{32.8}&
\underline{32.4}&
\textbf{29.0}&
\textbf{34.5} \\

\midrule

\multicolumn{1}{l}{
\textbf{Diffusion-based methods} } &
\multicolumn{16}{l}{
\textbf{One hypothesis selected} }\\ 
\midrule
(H=200, B) DiffPose~\cite{Holmquist_2023_ICCV} &
\textbf{28.1} &\underline{31.5} &\textbf{28.0} &\underline{30.8}& \underline{33.6}& 35.3& \textbf{28.5}& \textbf{27.6}& \underline{40.8}& 44.6& \underline{31.8}& \underline{32.1}& 32.6& \underline{28.1}& \underline{26.8} & \underline{32.0}\\

\multicolumn{17}{c}{\hdashrule[0.2ex]{17cm}{0.5pt}{1mm}} \\

(H=20, R) \textit{\method} & 34.4 & 37.7 & 35.1 & 38.1 & 36.9 & \underline{32.4} & 36.3 & 45.5 & 51.7 & \underline{37.2} & 42.0 & 34.6 & \underline{29.2} & 40.1 & 32.6 & 37.8 \\

(H=20, B) \textit{\method} & 
\underline{30.8}&
\textbf{29.2}&
\underline{29.0}&
\textbf{27.8}&
\textbf{28.2}&
\textbf{31.6}&
\underline{30.6}&
\underline{33.0}&
\textbf{32.1}&
\textbf{33.3}&
\textbf{29.6}&
\textbf{27.2}&
\textbf{28.3}&
\textbf{27.9}&
\textbf{24.5}&
\textbf{29.7} \\

\bottomrule
\end{tabular}
}
\end{table*}

\section{Datasets and metrics}

\subsection{Datasets}
\textbf{Human3.6M}
~\cite{ionescu2013human3}  is a large-scale dataset containing 3.6 million images with annotated 3D human poses with a skeleton of 17 joints. The dataset has been captured using four cameras from different views in an indoor environment. A total of $11$ subjects perform $15$ daily activities. To allow a fair comparison with the state-of-the-art, we train on subjects (S1, S5, S6, S7, S8), while we test on subjects (S9, S11). 
Furthermore, to underline the advantage of having a multi-hypothesis system 
and following the recent DiffPose \cite{Holmquist_2023_ICCV} work, we also tested {\method} on the \textbf{H36MA} subset of Human3.6M defined by~\cite{wehrbein2021probabilistic}. H36MA contains $6.4$\% of the test set, which includes highly ambiguous cases. 

\noindent \textbf{MPI-INF-3DHP}
this dataset~\cite{mehta2017monocular} has been captured with 14 cameras placed in both indoor and outdoor environments. More than 1.3 million frames have been captured and annotated using a skeleton model with 17 joints. The frames contain 8 actors performing 8 activities such as walking, sitting, sport, and dynamic actions. Following the official train/test split, we train our method using 8 camera views of the training set and evaluate the valid frames of the test set.

\subsection{Metrics} 
The Mean Per Joint Position Error (MPJPE)~\cite{joo_iccv_2015} calculates the average Euclidean distance between estimated joint 3D coordinates and their ground truth in millimeters is used. 

Moreover, on Human3.6M we also use the aligned version of the MPJPE (usually referred to P-MPJPE or reconstruction error \cite{hmrKanazawa17}), which is MPJPE after rigid alignment of the prediction with ground truth via Procrustes Analysis~\cite{Gower75}. On MPI-INF-3DHP the performance is assessed also in terms of the Percentage of Correct Keypoints (PCK) within a 150mm range and Area Under Curve (AUC) metrics.

\section{Experimental Results}

Taking inspiration from previous literature methods~\cite{shan2023diffusion}, we design the {\method} framework, adapting a transformer-based deterministic approach~\cite{zhao2023contextaware} to the diffusion model. 
In particular, the first attention module is a $4$-layer multi-head attention block that learns relations between context and pose features. The second attention module is similarly implemented in terms of learnable functions, but it learns correlations between the skeleton joints. Finally, a regression head projects the feature space from $d$ to $3$ dimensions. 

We train {\method} with a batch size of $128$ for $50$ epochs on both datasets. We use Adam optimizer~\cite{kingma2014adam} with a starting learning rate of $6e^{-4}$ using a linear decay with factor $0.993$. We apply random horizontal flipping as data augmentation for both images and skeletons. 
We implement {\method} using the PyTorch~\cite{paszke2019pytorch} framework and the \textit{Diffusers}~\footnote{\url{https://huggingface.co/docs/diffusers/index}} library for the diffusion-based technique. Moreover, to guarantee the full reproducibility of our results, the code and the training/testing hyperparameters will be publicly released. 

In this section, we focus on quantitative results, and refer the reader to the supplementary material for qualitative evaluations.

\clearpage

\begin{wraptable}[18]{r}{0.5\textwidth}
\centering
\caption{Results on the hard subset H36MA in terms of MPJPE and P-MPJPE. Number of hypotheses H and aggregation (\textbf{A}verage, \textbf{M}edian) or selection (\textbf{R}andom, \textbf{B}est) method are also reported (see Sect. \ref{subsec:aggregation}).}
\label{tab:human36MA}
\resizebox{\linewidth}{!}{
\begin{tabular}{l | c c }
\toprule
Method &  MPJPE & P-MPJPE   \\
&(mm) & (mm) \\ 
\midrule
Li \etal \cite{li2019generating}  & 81.1 & 66.0 \\
Sharma \etal \cite{sharma2019monocular}  & 78.3 & 61.1 \\
Wehrbein \etal \cite{wehrbein2021probabilistic}  & 71.0 & 54.2 \\
\midrule
(H=20, A) \textit{\method} & \underline{46.2} & \underline{36.5} \\
(H=20, M) \textit{\method} & \textbf{45.8} &\textbf{ 36.2} \\
\midrule
(H=200, B) DiffPose~\cite{Holmquist_2023_ICCV} & 63.1 & 46.7\\
(H=1, R) \textit{\method}& \underline{48.8} & \underline{39.8} \\
(H=20, B) \textit{\method} & \textbf{39.6} & \textbf{31.6} \\
\bottomrule
\end{tabular}
}
\end{wraptable}

\subsection{Comparison with state-of-the-art}

To make the comparison with state-of-the-art techniques more fair, we directly compared {\method} with single-frame approaches, able to work without the need for the temporal tracking of people. Thus, we focus on single-frame approaches, dividing diffusion-based methods from the others for clarity.

\subsubsection{Evaluation on Human3.60M.}
In Table~\ref{tab:mpjpe-human} and Table~\ref{tab:p-mpjpe-human}, we report the results in terms of MPJPE (Protocol 1) and P-MPJPE (Protocol 2) on the test set of Human3.6M. 
Following previous works that use CPN~\cite{chen2018cascaded} to get the 2D joints as input, we exploit the same network as the first step of our method.
We tested {\method} using all the selection and aggregation approaches described in Section \ref{subsec:aggregation}. 
As reported, {\method} surpasses its direct competitors~\cite{gong2023diffpose, choi2023diffupose, Holmquist_2023_ICCV} and achieves state-of-the-art performance across various actions and in the overall average score. Notably, the use of the median aggregation (\textbf{M} aggregation) yields even better results with respect to the average one (\textbf{A} aggregation), with a final average error of $42.8$mm using Protocol 1 and $34.5$mm using Protocol 2.

Noticeably, even when selecting a random hypothesis among the 20 generated ones (H=20, \textbf{R} selection), {\method} achieves high accuracy, only slightly inferior to that of the competitors. This result underscores the robustness and effectiveness of the proposed method, which generates plausible candidate poses.

As expected, the best results are obtained by selection with the ``Supervision from an Oracle''~\cite{sharma2019monocular} paradigm (\textbf{B} selection). However, as discussed above, the ground truth is normally not available in a real-world inference phase, and then it is not completely correct to use it to select the best hypothesis. 
This analysis is reported here for completeness and to better illustrate the potential theoretical capabilities of the proposed method in estimating the 3D position. This result reveals how a future study on aggregation techniques could further improve the results obtained with median-based aggregation.

The superior performance of SnapPose3D is also confirmed on the hardest subset of Human3.6, here referred to as H36MA~\cite{wehrbein2021probabilistic,Holmquist_2023_ICCV}, and reported in Table~\ref{tab:human36MA}. This evaluation can be considered an important result due to the ambiguity of the poses contained. In particular, our method outperforms
all comparable methods and in particular our direct competitor~\cite{Holmquist_2023_ICCV} by a large and consistent margin of $23.5$ and $15.1$mm for MPJPE and P-MPJPE, respectively.


\begin{table*}[t]
\centering
\caption{Results on MPI-INF-3DHP dataset. Following the common literature testing protocol, ground truth 2D poses are used as input.}
\label{tab:mpjpe_mpi}
\begin{tabular}{r c|c c c c c c c c c|c c c}
\toprule

\textbf{Method} &&& Pavvlo &  Zheng & Wang & Li & Zheng & Zhang  & Zhao 
&&& \textit{\method}  & \textit{\method} \\
 &&& \cite{pavllo20193d} &  \cite{zheng20213d} & \cite{wang2020motion} & \cite{li2022mhformer} & \cite{zheng20213d} & \cite{zhang2022mixste} & \cite{zhao2023contextaware} 
 &&& (H=20, R) & (H=20, M) \\

\midrule
\textbf{PCK $\uparrow$} &&& 86.0 & 88.6 & 86.9 & 93.8 & 95.4 & 94.4 & 96.8 
&&& 96.1 & \textbf{97.2} \\
\textbf{AUC $\uparrow$} &&& 51.9 & 56.4 & 62.1 & 63.3 & 63.2 & 66.5 & 70.7 
&&& 70.2 & \textbf{71.4} \\
\textbf{MPJPE $\downarrow$} &&& 84.0 & 77.1 & 68.1 & 58.0 & 57.7 & 54.9 & 44.7 
&&& 45.5 & \textbf{42.2} \\

\bottomrule
\end{tabular}
\end{table*}


\subsubsection{Evaluation on MPI-INF-3DHP.}
In Table~\ref{tab:mpjpe_mpi}, we report the results in terms of PCK, AUC, and MPJPE on the test set of MPI-INF-3DHP that, differently from the previous benchmark, contains outdoor sequences. 
Following the testing protocol of previous works~\cite{zheng20213d,li2022mhformer,zhang2022mixste}, we use the ground truth 2D pose detection as input. 
We trained the state-of-the-art competitor~\cite{zhao2023contextaware} on this dataset as well, using the publicly available implementation provided by the authors. 
Our approach achieves state-of-the-art results across all metrics, surpassing both non-temporal baselines and advanced temporal models, demonstrating strong generalization capabilities, especially in challenging outdoor environments.

\begin{wraptable}{r}{0.55\textwidth}
\centering
\caption{Ablation study on Human3.6M masking part of the conditioning input, reported through the MPJPE metric. 
}
    \label{tab:ablation_conditioning}
    \resizebox{\linewidth}{!}{
    \begin{tabular}{c c c|c c c c c}
    \toprule
     &  &&& \multicolumn{4}{c}{\textbf{Multiple Hypothesis Aggregation}} \\
    \textbf{Pose} & \textbf{Context} &&& \textbf{A} & \textbf{M} & \textbf{B} & \textbf{B} \\
     &  && &(avg) & (median)& (Pose-level) & (Joint-level) \\
    \midrule
    \checkmark &  &&& 59.9 & 59.8 & 56.8 & 49.9 \\
     & \checkmark &&& 45.3 & 45.4 & 42.3 & 37.6 \\
    \checkmark & \checkmark &&& \textbf{43.2} & \textbf{42.8} & \textbf{36.9} & \textbf{28.6} \\
    \bottomrule
    \end{tabular}
    }
\end{wraptable}

\subsubsection{Computational load and speed.}
Using batch size $B=1$, {\method} reaches an inference time ranging between $3$ and $10$ FPS considering up to $H=20$ hypotheses and  $K=20$ iterations. The GPU memory occupation is 708MB on a single Nvidia RTX 3090.
These results highlight how the system, although based on computationally demanding models (\textit{e.g.}, the diffusion-based model), can still achieve near real-time performance.

\subsection{Ablation studies}
\label{sec:ablation}

\subsubsection{Denoising conditioning.}
In Table~\ref{tab:ablation_conditioning}, we assess the performance of the diffusion-based denoising process depending on the visual feature conditioning. 
We report different MPJPE metrics aggregating or selecting the hypotheses with different techniques. It is worth noting that removing the context information from the learning process degrades the performance with -27.9\% on average while removing the pose features also impacts the results with -4.6\% on average. 
This analysis of the conditioning selection demonstrates that both visual and pose features are needed to obtain precise and reliable 3D human poses.

The last column reports the performance of selecting the closest joint to the ground truth (\textbf{B}est, joint-level) across all hypotheses. Since the best joints may come from different solutions, this represents a lower bound for selection approaches.

\begin{wrapfigure}{r}{0.39\textwidth}
    \centering
    \includegraphics[width=0.38\textwidth]{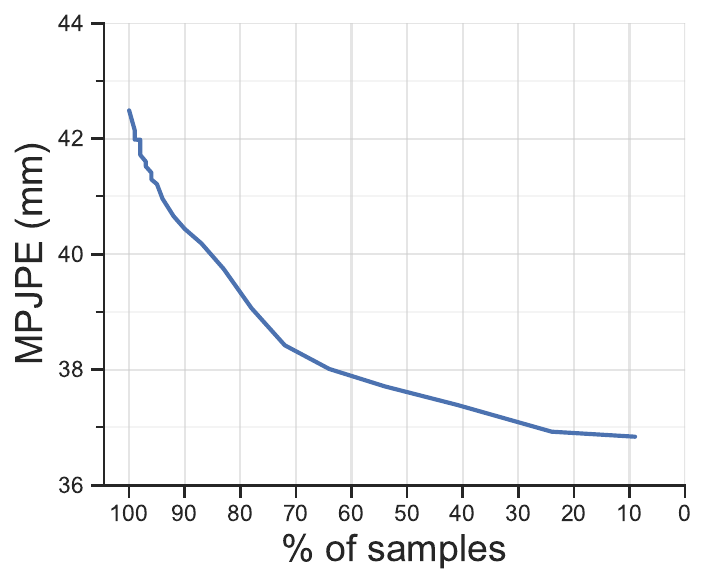}
    \caption{Confidence analysis in terms of MPJPE}
    \label{fig:ConfidenceAnalysis}
\end{wrapfigure}

\subsubsection{Confidence score.}
The generation of multiple and different but still plausible hypotheses not only allows for the improvement of performance thanks to aggregation techniques (see Sec. \ref{subsec:inference}), but also the definition of a confidence score.
To this aim, we computed and used as the confidence score of each pose prediction the variance between the H different provided hypotheses. More precisely, we computed the average between the variances at the joint coordinate level. 
For example, admitting a recall of around $90\%$ it is possible to improve the MPJPE metric by more than $2$mm.

To validate the chosen confidence score, we computed metrics on the most confident outputs, obtaining a reduction in global error at the cost of lower recall (i.e., reduced pose coverage), as shown in Figure~\ref{fig:ConfidenceAnalysis}. Notably, the confidence metric requires no ground truth, as it is computed directly on the generated outputs.

\begin{wrapfigure}{r}{0.39\textwidth}
    \centering
    \includegraphics[width=0.38\textwidth]{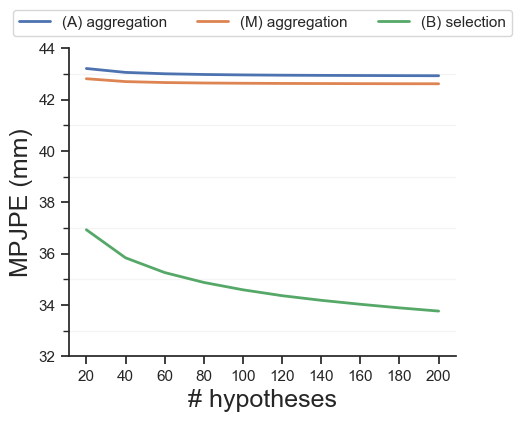}
    \caption{MPJPE vs. number of generated hypotheses}
    \label{fig:firstPlot}
\end{wrapfigure}

\subsubsection{Number of generated hypotheses.}
To evaluate the best number of hypotheses $H$ to generate, the MPJPE error (Protocol 1) was calculated with different aggregation/selection techniques on the Human3.6 dataset. The number of hypotheses varied between $20$ and $200$. 
The graph in Figure~\ref{fig:firstPlot} shows the results of {\method} using the aggregation methods proposed on the Human3.6 dataset. 
The median operator always outperforms the average one. The Oracle supervision generates the best solutions, as expected. However, median and average aggregations do not improve with more than 40 hypotheses, making $H=20$ the best compromise between performance and computational speed. 
On the contrary, the selection of oracles continues to improve as the number of hypotheses increases, as can be reasonably imagined from the generation of further possibilities from which to choose the best result.

\section{Conclusion}

In this work, we presented {\method}, a novel diffusion-based framework for 3D human pose estimation from a single image. Unlike most previous methods that rely on temporal sequences to resolve depth ambiguity, {\method} utilizes a single-frame approach, reducing the complexity of data acquisition and enabling online applications. 
By leveraging the generative capabilities of diffusion models, our method produces multiple plausible hypotheses for each input 2D pose and context feature.
Experimental results obtained on two public benchmarks confirm that the generation and aggregation of multiple hypotheses provide significant advantages.

\bibliographystyle{splncs04}
\bibliography{main}

@String(PAMI = {IEEE Trans. Pattern Anal. Mach. Intell.})

@String(CVPR= {IEEE Conf. Comput. Vis. Pattern Recog.})

@String(ICCV= {Int. Conf. Comput. Vis.})

@String(ECCV= {Eur. Conf. Comput. Vis.})

@String(ICPR = {Int. Conf. Pattern Recog.})

@String(ICLR = {Int. Conf. Learn. Represent.})

@String(PAMI  = {IEEE TPAMI})

@String(CVPR  = {CVPR})

@String(ICCV  = {ICCV})

@String(ECCV  = {ECCV})

@String(ICPR  = {ICPR})

@String(ICLR  = {ICLR})

@InProceedings{Holmquist_2023_ICCV,
    author    = {Holmquist, Karl and Wandt, Bastian},
    title     = {DiffPose: Multi-hypothesis Human Pose Estimation using Diffusion Models},
    booktitle = {ICCV},
    year      = {2023},
}

@article{bishop1994mixture,
    title        = {Mixture density networks},
    author       = {Bishop, Christopher M},
    year         = 1994,
    publisher    = {Aston University}
}

@inproceedings{cao2017realtime,
    title        = {Realtime multi-person 2d pose estimation using part affinity fields},
    author       = {Cao, Zhe and Simon, Tomas and Wei, Shih-En and Sheikh, Yaser},
    year         = 2017,
    booktitle    = CVPR,
    pages        = {7291--7299}
}

@inproceedings{chen2018cascaded,
    title        = {Cascaded pyramid network for multi-person pose estimation},
    author       = {Chen, Yilun and Wang, Zhicheng and Peng, Yuxiang and Zhang, Zhiqiang and Yu, Gang and Sun, Jian},
    year         = 2018,
    booktitle    = CVPR,
    pages        = {7103--7112}
}

@inproceedings{choi2023diffupose,
  author={Choi, Jeongjun and Shim, Dongseok and Kim, H. Jin},
  title={DiffuPose: Monocular 3D Human Pose Estimation via Denoising Diffusion Probabilistic Model}, 
  booktitle={2023 IEEE/RSJ International Conference on Intelligent Robots and Systems (IROS)}, 
  year={2023},
  pages={3773-3780}
}

@inproceedings{dhariwal2021diffusion,
    title        = {Diffusion models beat gans on image synthesis},
    author       = {Dhariwal, Prafulla and Nichol, Alexander},
    year         = 2021,
    booktitle    = "NeurIPS",
    volume       = 34,
    pages        = {8780--8794}
}

@inproceedings{gong2023diffpose,
    title        = {Diffpose: Toward more reliable 3d pose estimation},
    author       = {Gong, Jia and Foo, Lin Geng and Fan, Zhipeng and Ke, Qiuhong and Rahmani, Hossein and Liu, Jun},
    year         = 2023,
    booktitle    = CVPR,
    pages        = {13041--13051}
}

@article{Gower75,
    title = {Generalized procrustes analysis},
    author = {Gower, J.},
    year = {1975},
    journal = {Psychometrika},
    volume = {40},
    number = {1},
    pages = {33-51}
}

@article{ho2020denoising,
    title        = {Denoising diffusion probabilistic models},
    author       = {Ho, Jon and Jain, Ajay and Ab, Pieter},
    year         = 2020,
    journal      = "NeurIPS"
}

@inproceedings{hossain2018exploiting,
    title        = {Exploiting temporal information for 3d human pose estimation},
    author       = {Hossain, Mir Rayat Imtiaz and Little, James J},
    year         = 2018,
    booktitle    = ECCV,
    pages        = {68--84}
}

@article{ionescu2013human3,
    title        = {Human3.6m: Large scale datasets and predictive methods for 3d human sensing in natural environments},
    author       = {Ionescu, Catalin and Papava, Dragos and Olaru, Vlad and Sminchisescu, Cristian},
    year         = 2013,
    journal      = PAMI,
    volume       = 36,
    number       = 7,
    pages        = {1325--1339}
}

@inproceedings{jahangiri2017generating,
    title        = {Generating multiple diverse hypotheses for human 3d pose consistent with 2d joint detections},
    author       = {Jahangiri, Ehsan and Yuille, Alan L},
    year         = 2017,
    booktitle    = {ICCVW},
    pages        = {805--814}
}

@article{simoni2022semi,
  title={Semi-perspective decoupled heatmaps for 3d robot pose estimation from depth maps},
  author={Simoni, Alessandro and Pini, Stefano and Borghi, Guido and Vezzani, Roberto},
  journal={IEEE Robotics and Automation Letters},
  volume={7},
  number={4},
  year={2022},
  publisher={IEEE}
}

@article{simoni2024d,
  title={D-spdh: Improving 3d robot pose estimation in sim2real scenario via depth data},
  author={Simoni, Alessandro and Borghi, Guido and Garattoni, Lorenzo and Francesca, Gianpiero and Vezzani, Roberto},
  journal={IEEE Access},
  publisher={IEEE}
}

@inproceedings{joo_iccv_2015,
  title={Panoptic studio: A massively multiview system for social motion capture},
  author={Joo, Hanbyul and Liu, Hao and Tan, Lei and Gui, Lin and Nabbe, Bart and Matthews, Iain and Kanade, Takeo and Nobuhara, Shohei and Sheikh, Yaser},
  booktitle=ICCV,
  pages={3334--3342},
  year={2015}
}

@inproceedings{hmrKanazawa17,
  title={End-to-end Recovery of Human Shape and Pose},
  author = {Angjoo Kanazawa and Michael J. Black and David W. Jacobs and Jitendra Malik},
  booktitle=CVPR,
  year={2018}
}

@inproceedings{d2021refinet,
  title={Refinet: 3d human pose refinement with depth maps},
  author={D'Eusanio, Andrea and Pini, Stefano and Borghi, Guido and Vezzani, Roberto and Cucchiara, Rita},
  booktitle={2020 25th International Conference on Pattern Recognition (ICPR)},
  pages={2320--2327},
  year={2021},
  organization={IEEE}
}

@article{kingma2014adam,
    title        = {Adam: A method for stochastic optimization},
    author       = {Kingma, Diederik P and Ba, Jimmy},
    year         = 2014,
    journal      = {arXiv preprint arXiv:1412.6980}
}

@inproceedings{kipf2016semi,
    title        = {Semi-supervised classification with graph convolutional networks},
    author       = {Kipf, Thomas N and Welling, Max},
    year         = 2016,
    booktitle      = ICLR
}

@article{d2023depth,
  title={Depth-based 3D human pose refinement: Evaluating the refinet framework},
  author={D’Eusanio, Andrea and Simoni, Alessandro and Pini, Stefano and Borghi, Guido and Vezzani, Roberto and Cucchiara, Rita},
  journal={Pattern Recognition Letters},
  volume={171},
  pages={185--191},
  year={2023},
  publisher={Elsevier}
}

@inproceedings{li2019generating,
    title        = {Generating multiple hypotheses for 3d human pose estimation with mixture density network},
    author       = {Li, Chen and Lee, Gim Hee},
    year         = 2019,
    booktitle    = CVPR,
    pages        = {9887--9895}
}

@inproceedings{li2022mhformer,
    title        = {Mhformer: Multi-hypothesis transformer for 3d human pose estimation},
    author       = {Li, Wenhao and Liu, Hong and Tang, Hao and Wang, Pichao and Van Gool, Luc},
    year         = 2022,
    booktitle    = CVPR,
    pages        = {13147--13156}
}

@inproceedings{liu2020comprehensive,
    title        = {A comprehensive study of weight sharing in graph networks for 3d human pose estimation},
    author       = {Liu, Kenkun and Ding, Rongqi and Zou, Zhiming and Wang, Le and Tang, Wei},
    year         = 2020,
    booktitle    = ECCV
}

@inproceedings{martinez2017simple,
    title        = {A simple yet effective baseline for 3d human pose estimation},
    author       = {Martinez, Julieta and Hossain, Rayat and Romero, Javier and Little, James J},
    year         = 2017,
    booktitle    = ICCV,
    pages        = {2640--2649}
}

@inproceedings{mehta2017monocular,
    title        = {Monocular 3d human pose estimation in the wild using improved cnn supervision},
    author       = {Mehta, Dushyant and Rhodin, Helge and Casas, Dan and Fua, Pascal and Sotnychenko, Oleksandr and Xu, Weipeng and Theobalt, Christian},
    year         = 2017,
    booktitle    = {3DV},
    pages        = {506--516}
}

@inproceedings{oikarinen2021graphmdn,
    title        = {GraphMDN: Leveraging graph structure and deep learning to solve inverse problems},
    author       = {Oikarinen, Tuomas and Hannah, Daniel and Kazerounian, Sohrob},
    year         = 2021,
    booktitle    = {IJCNN},
    pages        = {1--9}
}

@inproceedings{pavlakos2017coarse,
    title        = {Coarse-to-fine volumetric prediction for single-image 3D human pose},
    author       = {Pavlakos, Georgios and Zhou, Xiaowei and Derpanis, Konstantinos G and Daniilidis, Kostas},
    year         = 2017,
    booktitle    = CVPR,
    pages        = {7025--7034}
}

@inproceedings{rombach2022high,
    title        = {High-resolution image synthesis with latent diffusion models},
    author       = {Rombach, Robin and Blattmann, Andreas and Lorenz, Dominik and Esser, Patrick and Ommer, Bj{\"o}rn},
    year         = 2022,
    booktitle    = CVPR,
    pages        = {10684--10695}
}

@inproceedings{shan2023diffusion,
  title={Diffusion-Based 3D Human Pose Estimation with Multi-Hypothesis Aggregation},
  author={Shan, Wenkang and Liu, Zhenhua and Zhang, Xinfeng and Wang, Zhao and Han, Kai and Wang, Shanshe and Ma, Siwei and Gao, Wen},
  booktitle=ICCV,
  year={2023}
}

@inproceedings{sharma2019monocular,
    title        = {Monocular 3d human pose estimation by generation and ordinal ranking},
    author       = {Sharma, Saurabh and Varigonda, Pavan Teja and Bindal, Prashast and Sharma, Abhishek and Jain, Arjun},
    year         = 2019,
    booktitle    = ICCV
}

@inproceedings{simo2012single,
    title        = {Single image 3D human pose estimation from noisy observations},
    author       = {Simo-Serra, Edgar and Ramisa, Arnau and Alenya, Guillem and Torras, Carme and Moreno-Noguer, Francesc},
    year         = 2012,
    booktitle    = CVPR
}

@inproceedings{sminchisescu2003kinematic,
    title        = {Kinematic jump processes for monocular 3D human tracking},
    author       = {Sminchisescu, Cristian and Triggs, Bill},
    year         = 2003,
    booktitle    = CVPR
}

@inproceedings{sohl2015deep,
    title        = {Deep unsupervised learning using nonequilibrium thermodynamics},
    author       = {Sohl-Dickstein, Jascha and Weiss, Eric and Maheswaranathan, Niru and Ganguli, Surya},
    year         = 2015,
    booktitle    = "ICML"
}

@inproceedings{song2021denoising,
    title        = {Denoising Diffusion Implicit Models},
    author       = {Jiaming Song and Chenlin Meng and Stefano Ermon},
    year         = 2021,
    booktitle    = ICLR
}

@inproceedings{sun2019deep,
    title        = {Deep high-resolution representation learning for human pose estimation},
    author       = {Sun, Ke and Xiao, Bin and Liu, Dong and Wang, Jingdong},
    year         = 2019,
    booktitle    = CVPR,
    pages        = {5693--5703}
}

@inproceedings{sun2017compositional,
    title        = {Compositional human pose regression},
    author       = {Sun, Xiao and Shang, Jiaxiang and Liang, Shuang and Wei, Yichen},
    year         = 2017,
    booktitle    = ICCV,
    pages        = {2602--2611}
}

@inproceedings{wang2020motion,
    title        = {Motion guided 3d pose estimation from videos},
    author       = {Wang, Jingbo and Yan, Sijie and Xiong, Yuanjun and Lin, Dahua},
    year         = 2020,
    booktitle    = ECCV,
    pages        = {764--780}
}

@inproceedings{wehrbein2021probabilistic,
	title        = {Probabilistic monocular 3d human pose estimation with normalizing flows},
	author       = {Wehrbein, Tom and Rudolph, Marco and Rosenhahn, Bodo and Wandt, Bastian},
	year         = 2021,
	booktitle    = ICCV,
	pages        = {11199--11208}
}

@inproceedings{xu2021graph,
    title        = {Graph stacked hourglass networks for 3d human pose estimation},
    author       = {Xu, Tianhan and Takano, Wataru},
    year         = 2021,
    booktitle    = CVPR,
    pages        = {16105--16114}
}

@inproceedings{yang20183d,
    title        = {3d human pose estimation in the wild by adversarial learning},
    author       = {Yang, Wei and Ouyang, Wanli and Wang, Xiaolong and Ren, Jimmy and Li, Hongsheng and Wang, Xiaogang},
    year         = 2018,
    booktitle    = CVPR,
    pages        = {5255--5264}
}

@inproceedings{zhang2022mixste,
    title        = {MixSTE: Seq2seq Mixed Spatio-Temporal Encoder for 3D Human Pose Estimation in Video},
    author       = {Zhang, Jinlu and Tu, Zhigang and Yang, Jianyu and Chen, Yujin and Yuan, Junsong},
    year         = 2022,
    booktitle    = CVPR
}

@inproceedings{zhaoCVPR19semantic,
    title        = {Semantic Graph Convolutional Networks for 3D Human Pose Regression},
    author       = {Zhao, Long and Peng, Xi and Tian, Yu and Kapadia, Mubbasir and Metaxas, Dimitris N.},
    year         = 2019,
    booktitle    = CVPR,
    pages        = {3425--3435}
}

@inproceedings{
    zhao2023contextaware,
    title={A Single 2D Pose with Context is Worth Hundreds for 3D Human Pose Estimation},
    author={Zhao, Qitao and Zheng, Ce and Liu, Mengyuan and Chen, Chen},
    booktitle="NeurIPS",
    year={2023},
}

@inproceedings{zhao2022graformer,
    title        = {GraFormer: Graph-Oriented Transformer for 3D Pose Estimation},
    author       = {Zhao, Weixi and Wang, Weiqiang and Tian, Yunjie},
    year         = 2022,
    booktitle    = CVPR,
    pages        = {20438--20447}
}

@inproceedings{pavllo20193d,
    title        = {3d human pose estimation in video with temporal convolutions and semi-supervised training},
    author       = {Pavllo, Dario and Feichtenhofer, Christoph and Grangier, David and Auli, Michael},
    year         = 2019,
    booktitle    = CVPR
}

@inproceedings{zheng20213d,
    title        = {3d human pose estimation with spatial and temporal transformers},
    author       = {Zheng, Ce and Zhu, Sijie and Mendieta, Matias and Yang, Taojiannan and Chen, Chen and Ding, Zhengming},
    year         = 2021,
    booktitle    = ICCV
}

@article{paszke2019pytorch,
  title={Pytorch: An imperative style, high-performance deep learning library},
  author={Paszke, Adam and Gross, Sam and Massa, Francisco and Lerer, Adam and Bradbury, James and Chanan, Gregory and Killeen, Trevor and Lin, Zeming and Gimelshein, Natalia and Antiga, Luca and others},
  journal="NeurIPS",
  year={2019}
}

@misc{diffusers,
  title = {Diffusers Library},
  howpublished={\url{https://huggingface.co/docs/diffusers/index}}
}

@inproceedings{von2018recovering,
  title={Recovering accurate 3d human pose in the wild using imus and a moving camera},
  author={Von Marcard, Timo and Henschel, Roberto and Black, Michael J and Rosenhahn, Bodo and Pons-Moll, Gerard},
  booktitle=ECCV,
  pages={601--617},
  year={2018}
}

@article{vaswani2017attention,
  title={Attention is all you need},
  author={Vaswani, A},
  journal={NeurIPS},
  year={2017}
}

@article{terreran2023general,
  title={A general skeleton-based action and gesture recognition framework for human--robot collaboration},
  author={Terreran, Matteo and Barcellona, Leonardo and Ghidoni, Stefano},
  journal={Robotics and Autonomous Systems},
  volume={170},
  pages={104523},
  year={2023},
  publisher={Elsevier}
}

@article{arac2019deepbehavior,
  title={DeepBehavior: A deep learning toolbox for automated analysis of animal and human behavior imaging data},
  author={Arac, Ahmet and Zhao, Pingping and Dobkin, Bruce H and Carmichael, S Thomas and Golshani, Peyman},
  journal={Frontiers in systems neuroscience},
  volume={13},
  pages={20},
  year={2019},
  publisher={Frontiers Media SA}
}

@inproceedings{liu2022ego+,
  title={Ego+ x: An egocentric vision system for global 3d human pose estimation and social interaction characterization},
  author={Liu, Yuxuan and Yang, Jianxin and Gu, Xiao and Guo, Yao and Yang, Guang-Zhong},
  booktitle={2022 IEEE/RSJ International Conference on Intelligent Robots and Systems (IROS)},
  pages={5271--5277},
  year={2022},
  organization={IEEE}
}

@article{8765346,
  author = {Z. {Cao} and G. {Hidalgo Martinez} and T. {Simon} and S. {Wei} and Y. A. {Sheikh}},
  journal = {IEEE Transactions on Pattern Analysis and Machine Intelligence},
  title = {OpenPose: Realtime Multi-Person 2D Pose Estimation using Part Affinity Fields},
  year = {2019}
}

@article{xu2022vitpose,
  title={Vitpose: Simple vision transformer baselines for human pose estimation},
  author={Xu, Yufei and Zhang, Jing and Zhang, Qiming and Tao, Dacheng},
  journal={Adv. in neural information proces. systems},
  volume={35},
  year={2022}
}

@inproceedings{li2019crowdpose,
  title={Crowdpose: Efficient crowded scenes pose estimation and a new benchmark},
  author={Li, Jiefeng and Wang, Can and Zhu, Hao and Mao, Yihuan and Fang, Hao-Shu and Lu, Cewu},
  booktitle={Proceedings of the IEEE/CVF conference on computer vision and pattern recognition},
  year={2019}
}

@inproceedings{lin2014microsoft,
  title={Microsoft coco: Common objects in context},
  author={Lin, Tsung-Yi and Maire, Michael and Belongie, Serge and Hays, James and Perona, Pietro and Ramanan, Deva and Doll{\'a}r, Piotr and Zitnick, C Lawrence},
  booktitle={Computer vision--ECCV 2014: 13th European conference, zurich, Switzerland, September 6-12, 2014, proceedings, part v 13},
  pages={740--755},
  year={2014},
  organization={Springer}
}

@article{wang2021deep,
  title={Deep 3D human pose estimation: A review},
  author={Wang, Jinbao and Tan, Shujie and Zhen, Xiantong and Xu, Shuo and Zheng, Feng and He, Zhenyu and Shao, Ling},
  journal={Computer Vision and Image Understanding},
  volume={210},
  year={2021},
  publisher={Elsevier}
}

@article{paudel2022industrial,
  title={Industrial ergonomics risk analysis based on 3d-human pose estimation},
  author={Paudel, Prabesh and Kwon, Young-Jin and Kim, Do-Hyun and Choi, Kyoung-Ho},
  journal={Electronics},
  volume={11},
  number={20},
  pages={3403},
  year={2022},
  publisher={MDPI}
}

@article{fan20243d,
  title={3D pose estimation dataset and deep learning-based ergonomic risk assessment in construction},
  author={Fan, Chao and Mei, Qipei and Li, Xinming},
  journal={Automation in Construction},
  volume={164},
  year={2024},
  publisher={Elsevier}
}

@article{sarafianos20163d,
  title={3d human pose estimation: A review of the literature and analysis of covariates},
  author={Sarafianos, Nikolaos and Boteanu, Bogdan and Ionescu, Bogdan and Kakadiaris, Ioannis A},
  journal={Computer Vision and Image Understanding},
  volume={152},
  pages={1--20},
  year={2016},
  publisher={Elsevier}
}

\end{document}


\title{Supplementary Material \\
{\method}: Diffusion-Based Single-Frame \\ 2D-to-3D Lifting of Human Poses}

\titlerunning{Diffusion-Based Single-Frame 2D-to-3D Lifting of Human Poses}
%
\author{
Alessandro~Simoni\inst{1} \and
Riccardo~Catalini\inst{1} \and
Davide~Di~Nucci\inst{1} \and
Guido~Borghi\inst{1} \and
Davide~Davoli\inst{2}\thanks{Providing contracted services} \and
Lorenzo~Garattoni\inst{2} \and
Gianpiero~Francesca\inst{2} \and
Yuki~Kawana\inst{3} \and
Roberto~Vezzani\inst{1}
}


\institute{
University of Modena and Reggio Emilia (UniMoRe), Italy \and
Toyota Motor Europe (TME), Brussels, Belgium \and
Woven by Toyota, Japan
}

%
%
%
\maketitle              

\section{Qualitative results}
\begin{figure*}[t!]
    \centering
    \includegraphics[width=\linewidth]{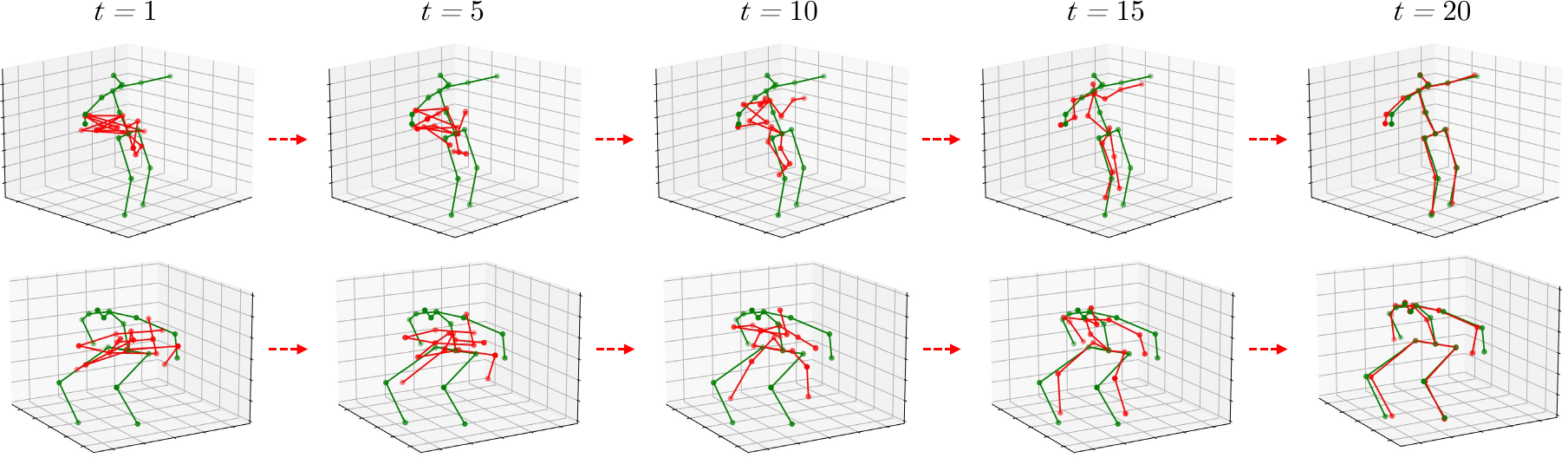}
    \caption{Visualization of the denoising process of {\method} during different iterations with $K=20$ on two sample frames. The green skeletons are the ground truth, and the red skeletons are predicted by our method using the median Aggregation on H=20 hypotheses.}
    \label{fig:qualitative_iterations}
\end{figure*}

To better underline the behavior of the framework during each denoising step, Figure \ref{fig:qualitative_iterations} reports the output pose provided after $t$ iterations out of $T=20$ for two random test examples. As shown, the noise is gradually removed until the final step while, after the first iteration, the output pose is still completely random. 

\begin{figure*}[th!]
    \centering
    \includegraphics[width=\linewidth]{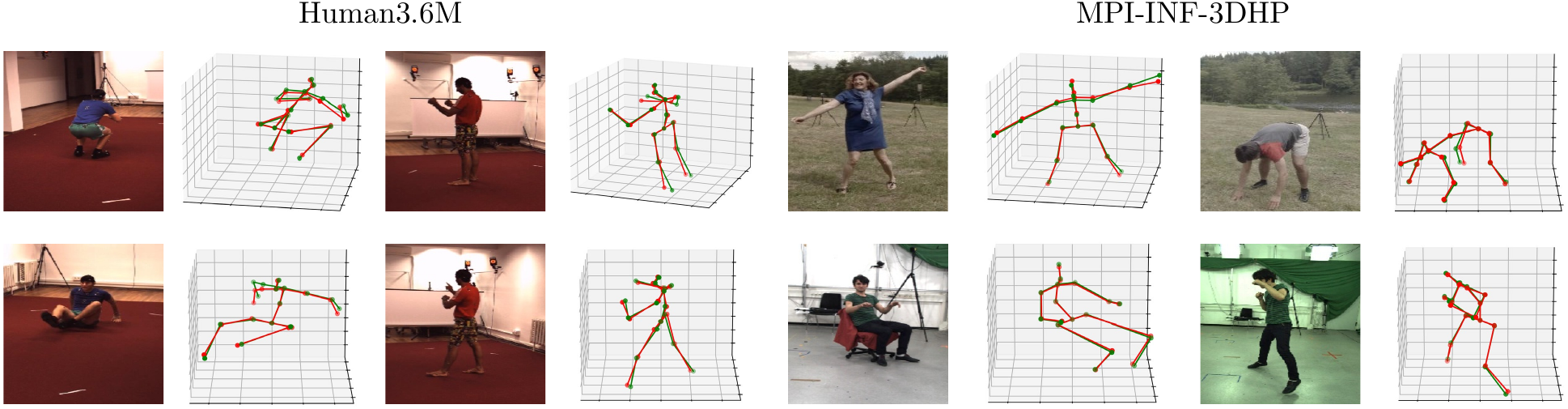}
    \caption{Qualitative results of {\method} on Human3.6M and MPI-INF-3DHP. The green skeletons are the ground truth, and the red skeletons are the predicted pose output by SnapPose3D.}
    \label{fig:qualitative_results}
\end{figure*}

\begin{figure*}
    \centering
    \includegraphics[width=0.75\linewidth]{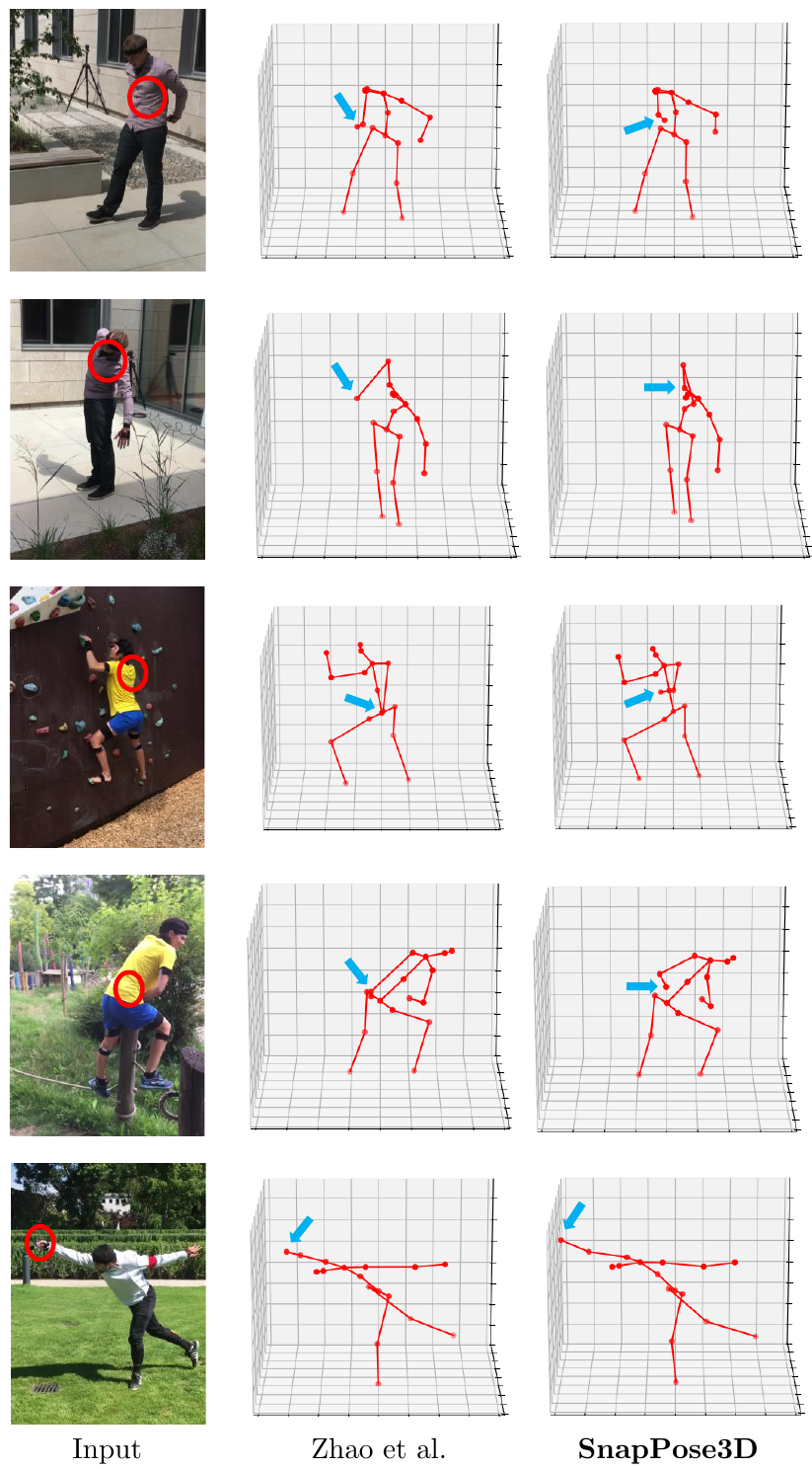}
    \caption{Qualitative comparison between {\method} and Zhao \etal~\cite{zhao2023contextaware} on in-the-wild videos of 3DPW dataset~\cite{von2018recovering}.}
    \label{fig:3dpw-comparison}
\end{figure*}

Some qualitative results obtained by SnapPose3D on the Human3.6 dataset are reported in Figure \ref{fig:qualitative_results}.
In Figure~\ref{fig:3dpw-comparison}, we qualitatively compare {\method} with Zhao et al.~\cite{zhao2023contextaware} on the 3DPW~\cite{von2018recovering} dataset, using blue arrows to highlight the robustness of our predictions in cases of occlusions. Notably, the multi-hypothesis approach of {\method} proves beneficial for obtaining more reliable human poses under challenging conditions.

\section{Additional qualitative results}
\begin{figure*}
    \centering
    \includegraphics[width=0.75\linewidth]{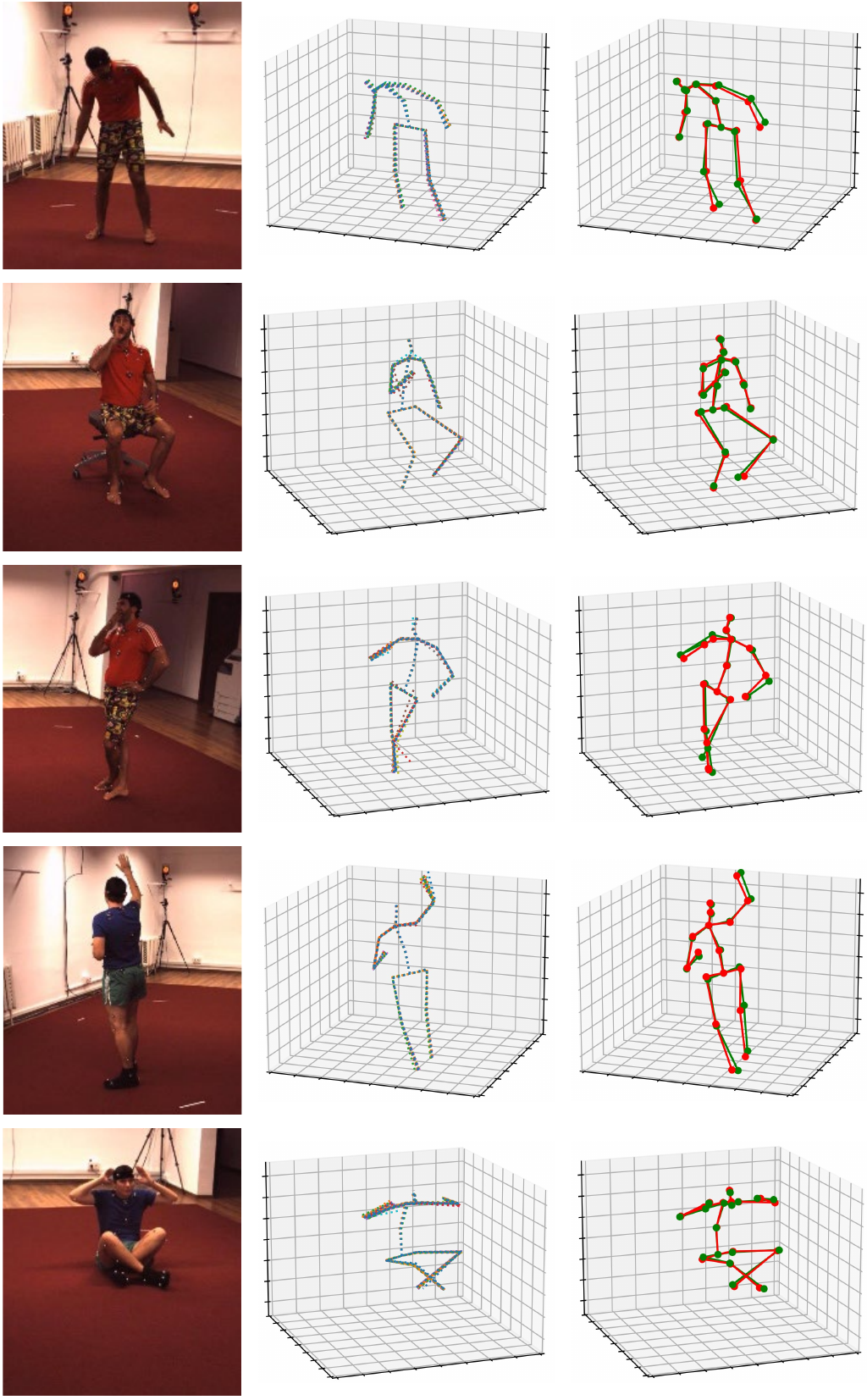}
    \caption{Qualitative results on Human3.6M. From left to right it shows the input image, the predicted hypotheses (dashed colored skeletons), and the average pose with regard to the ground truth (red and green skeletons respectively).}
    \label{fig:qualitative-h36m}
\end{figure*}

\begin{figure*}
    \centering
    \includegraphics[width=0.75\linewidth]{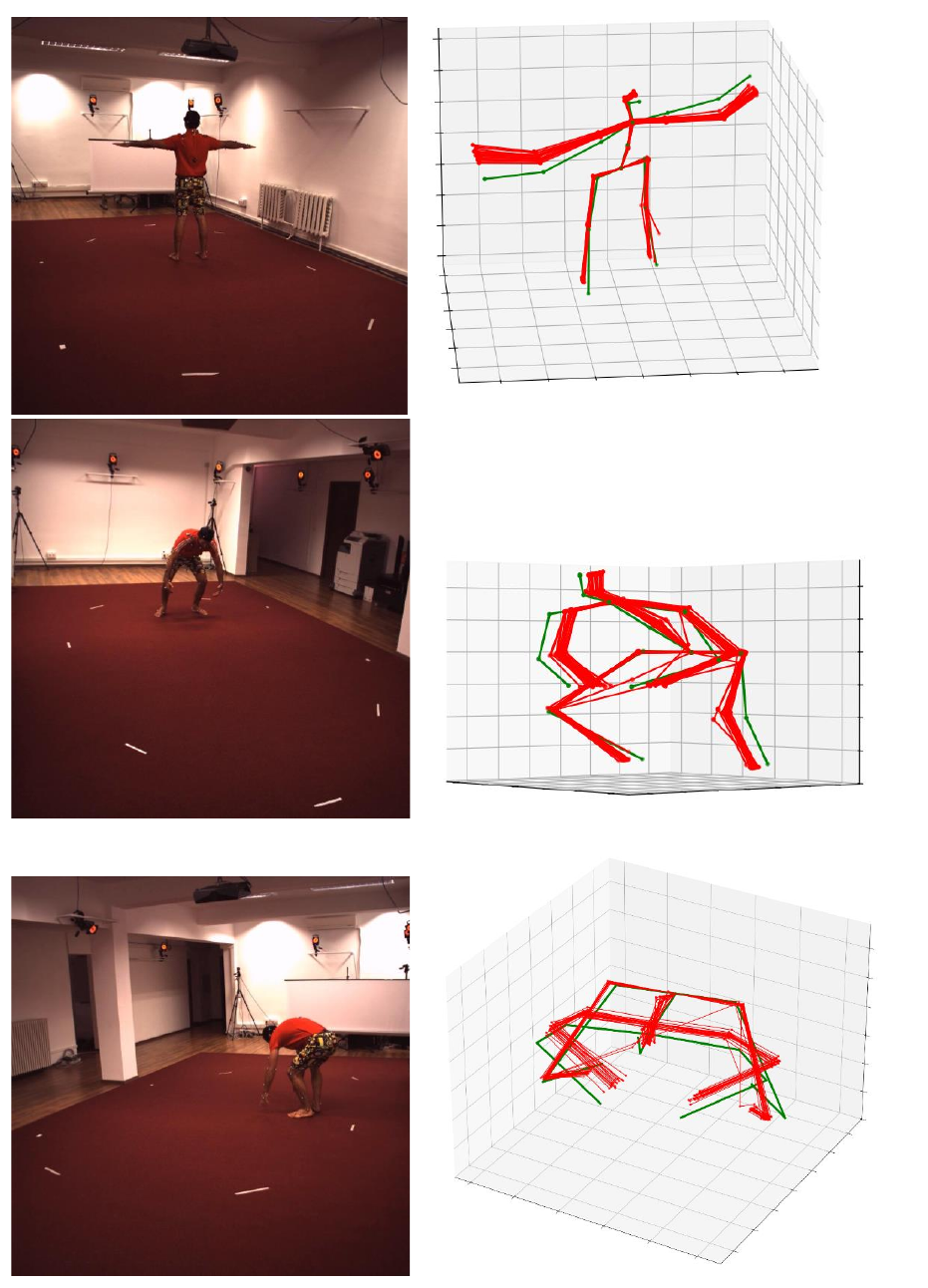}
    \caption{Qualitative results on Human3.6 dataset (with some frames of the hard subset H36MA) with the 20 hypotheses generated by {\method} in red. The ground truth is reported in green.}
    \label{fig:qualitative-h36m-2}
\end{figure*}

\begin{figure*}
    \centering
    \includegraphics[width=0.75\linewidth]{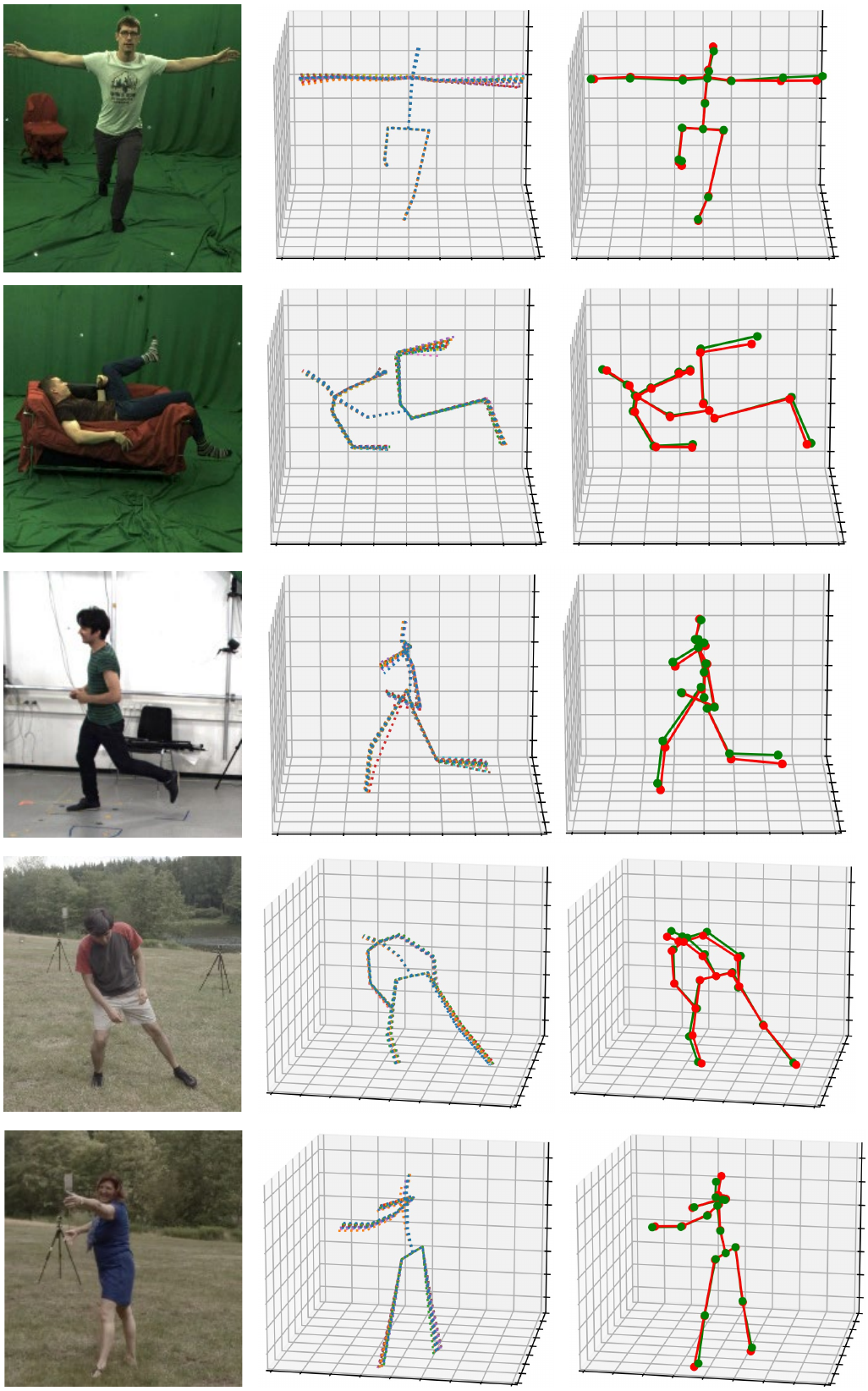}
    \caption{Qualitative results on MPI-INF-3DHP. From left to right it shows the input image, the predicted hypotheses (dashed colored skeletons), and the average pose with regard to the ground truth (red and green skeletons respectively).}
    \label{fig:qualitative-mpi}
\end{figure*}

In Figure~\ref{fig:qualitative-h36m}, ~\ref{fig:qualitative-h36m-2}, and Figure~\ref{fig:qualitative-mpi}, additional qualitative results show the average pose with regard to the ground truth along with all the 20 hypotheses solutions of {\method}. It is worth noting that the diversity in the hypothesis generation is correlated to the joints that are moving in the actions, \ie arms or legs.

\section{Reproducibility}
We implement {\method} using the PyTorch~\cite{paszke2019pytorch} framework and the \textit{Diffusers}~\cite{diffusers} library for the diffusion-based technique. Moreover, to guarantee the full reproducibility of our results, we will release upon paper acceptance the code and the training/testing hyperparameters.

\bibliographystyle{splncs04}
\bibliography{main}